# A Deep CNN Model for Ringing Effect Attenuation of Vibroseis Data

*Zhuang Jia, Wenkai Lu,*
*Tsinghua University, Beijing, P.R.China*

**Summary**

In the field of exploration geophysics, seismic vibrator is one of the widely used seismic sources to acquire seismic data, which is usually named vibroseis. "Ringing effect" is a common problem in vibroseis data processing due to the limited frequency bandwidth of the vibrator, which degrades the performance of first-break picking. In this paper, we proposed a novel deringing model for vibroseis data using deep convolutional neural network (CNN). In this model we use end-to-end training strategy to obtain the deringed data directly, and skip connections to improve model training process and preserve the details of vibroseis data. For real vibroseis deringing task we synthesize training data and corresponding labels from real vibroseis data and utilize them to train the deep CNN model. Experiments are conducted both on synthetic data and real vibroseis data. The experiment results show that deep CNN model can attenuate the ringing effect effectively and expand the bandwidth of vibroseis data. The STA/LTA ratio method for first-break picking also shows improvement on deringed vibroseis data using deep CNN model.

**Introduction**

Vibroseis technology has gained the most popularity in the field of land seismic exploration [1]. Land vibroseis has shown to be low-cost and have the possibility to separate each source, which makes it suitable for simultaneous sourcing [2]. An important component of vibroseis technique is the seismic vibrators. Vibrators are used as sources to generate energy signals which propagate into underground over a period of time. The recorded signals can be used for interpretation after correlated with reference signals. Different from the impulsive source (e.g. dynamite) which can be assumed as near instantaneous spike (thus has a very broad frequency bandwidth), the vibrator signal has a limited frequency bandwidth. As a result, the vibroseis data acquired has obvious "ringing effect", which makes it hard to pick the right first-break of each trace, and thus interferes the subsequent processes.

The difficulty of ringing suppression task for vibroseis data is to restore the missing frequency spectrum to a certain extent, thus broaden the frequency bandwidth to attenuate the influence caused by ringing effect in the subsequent processing steps. Various approaches has been brought up in order to improve the quality of vibroseis data. One of the common methods is deconvolution. Deconvolution is actually a kind of frequency filtering with the form of convolution, and the objective of it is to compress the seismic wavelet and improve the resolution. A number of methods using deconvolution have been proposed: the reweighting strategies for deconvolution are used in [3] as a regularization term to diminish the influence of the outliers. In [4] Monte Carlo Markov Chain (MCMC) method is applied to estimate the Gaussian mixture parameters which is used in deconvolution. The Multichannel Semi-blind Deconvolution model (MSBD) is used to recover both reflectivity and wavelet [5] . Frequency-domain sweep deconvolution (FDSD) uses frequency domain to simplify the convolution to multiplication [6]. Besides deconvolution based methods, wavelet transform has also been used to improve the quality of seismic data and enhance the first-break [7].

In the area of natural image processing, convolutional neural network (CNN) has been widely and successfully applied to many low-level vision tasks, such as denoising [8], deblocking and deringing for compressed images [9], super resolution [10], showing that CNN networks have the ability to learn these low-level transformations between train data and reference label. With the development of network training techniques, CNNs with deeper structure (more layers) tend to show stronger ability to extract the information of the transform between train data and reference label. Inspired by the deep CNN models in natural image processing tasks, we proposed a deep CNN model for vibroseis deringing task, and synthesized the train data and reference label using real vibroseis data. After the training finished, we applied the trained deep CNN network to the whole vibroseis data. Applications on both synthetic and real vibroseis data demonstrate that deep CNN model can reach promising results.

**Deep CNN network for vibroseis deringing**

Our deep convolutional neural network model for ringing effect attenuation of vibroseis is shown in Figure 1. There are 9 layers in this deep CNN model with two skip connections. The main components of this network are convolutional operation (conv), batch normalization operation (BN), and activation function (ReLU or tanh). Convolution is used to extract the features and details of the input data with less parameters by local perception and weight sharing. Batch normalization is a common strategy applied in training deep neural network, which can reduce the internal covariate shift problem. Activation provides non-linearity of neural network and enhances the ability to learn the mappings from input to output. All the layers in the deep CNN network have the same structure of

A Deep CNN Model for Ringing Effect Attenuation of Vibroseis Data

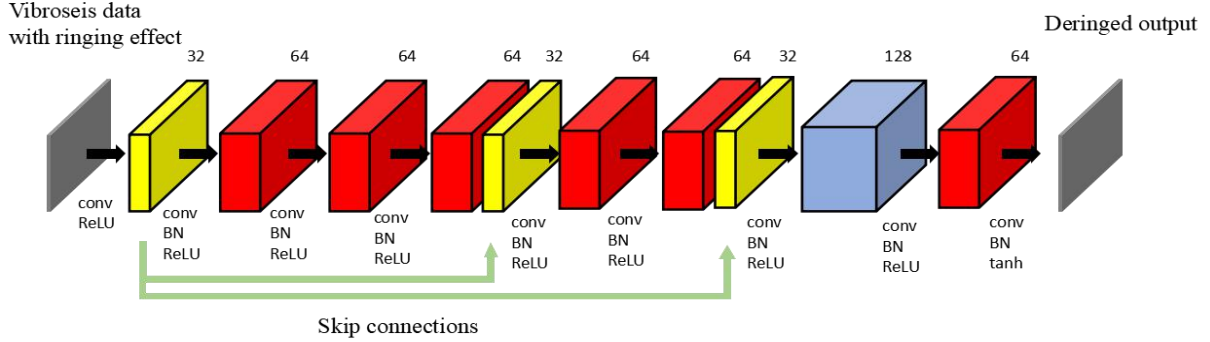

Figure 1: Deep CNN model architecture

conv+BN+ReLU except for the first and last layer. The first layer converts the input vibroseis data with ringing effect into 32 feature maps. As our input data has been already normalized, so the BN operation is not necessary. The last layer outputs our deringed result, which is expected to be in range -1 to 1 (normalized). So we use tanh activation instead of ReLU in the last layer. The input layer and the hidden layers of the deep CNN model can be expressed as below:

$$o_{1,c} = ReLU(k_{1,c} * x + b_{1,c}), c = 1,\ldots,C \qquad (1)$$

$$o_{j,c} = ReLU(k_{j,c} * n_j + b_{j,c}),$$
$$j = 2,\ldots,L-1; c = 1,\ldots,C_j \qquad (2)$$

where $x$ represents the input vibroseis data with ringing effect, $n_j$ represents the input of the $j$ th layer. $k_{j,c}$ and $b_{j,c}$ are the $c$ th convolutional kernel and bias in the $j$ th layer respectively. $o_{j,c}$ is the feature map of the $c$ th kernel of the $j$ th layer, $C_j$ is the number of kernels in the $j$ th layer, and $L$ is the total number of layers of the deep CNN model. Using the same form, the last layer can be expressed as :

$$\tilde{x} = \tanh(BN(k_L * n_L + b_L)) \qquad (3)$$

In the training process, the training data $s$ and label $\hat{s}$ are synthesized from real vibroseis data $y$. The objective is to minimize the mean-square error (MSE) of deep CNN output $\tilde{x}$ and label $\hat{x}$ using back propagation:

$$\min_{k,b} \| \tilde{s} - \hat{s} \|^2 \qquad (4)$$

After training process finished, real data $y$ is used as the input of the deep CNN model, by forward propagation, output $\tilde{y}$ is calculated as the deringed vibroseis data.

**Experiments on synthetic and real seismic data**

In order to explain that our deep CNN model can learn the mapping of ringing seismic data to deringed seismic data,

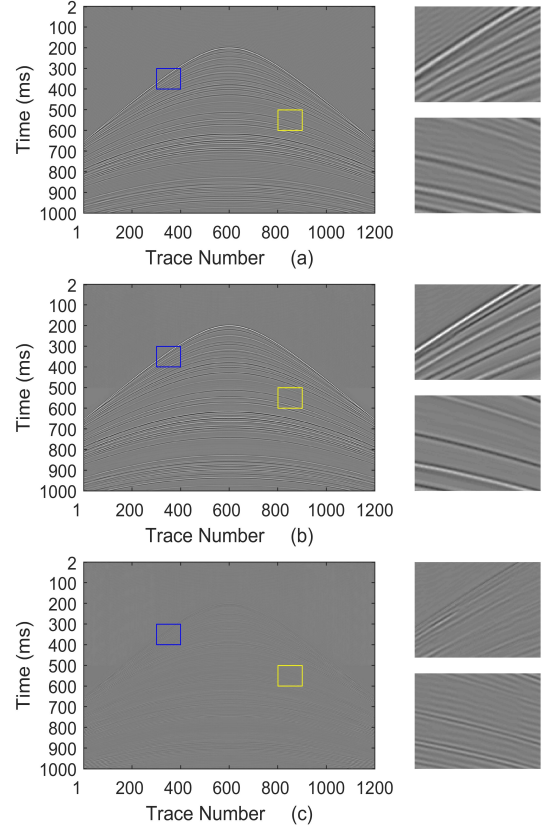

Figure 2: Comparison of ringing synthetic seismic data before and after deep CNN model. (a) is synthetic ringing data as network input, (b) is model deringed result. The right side of each sub-figure is the enlarged display of the blue and yellow frames labeled regions from top to bottom respectively. (c) shows the difference of deringed result and original synthetic data without band-pass filtering.

- 2 -

A Deep CNN Model for Ringing Effect Attenuation of Vibroseis Data

we apply the deep CNN model to synthetic seismic data. The parameters of our synthetic seismic data is as follows: number of time samples is 1000, number of space samples is 1200, time sampling rate is 500 Hz, spatial sampling interval is 3.125 m, velocities are from 1300 to 2300 m/s, and we use Ricker wavelet with dominant frequency of 60 Hz as our seismic wavelet. This synthetic seismic data is used as the train label without ringing effect. To generate corresponding ringing data, we filter the synthetic seismic data using a band-pass filter with passband from 6 Hz to 72 Hz. After ideal band-pass filtering method, there are obvious ringing effect in the synthetic seismic data. We use band-pass filtered data with ringing effect as our train data, and the corresponding original data (before band-pass filtering) as reference labels, and train the deep CNN model for 20 epochs, we apply the trained deep CNN model to a different synthetic seismic data filtered by the same ideal band-pass filter. The result of this experiment is shown in Figure 2. In addition, Figure 3 is the comparison of input ringing data, deep CNN deringed data, and original data in the f-K domain. It can be seen that the missing frequencies are restored to a certain extent. Experiments on synthetic seismic data illustrate that deep CNN model can attenuate ringing and broaden the bandwidth of synthetic seismic data.

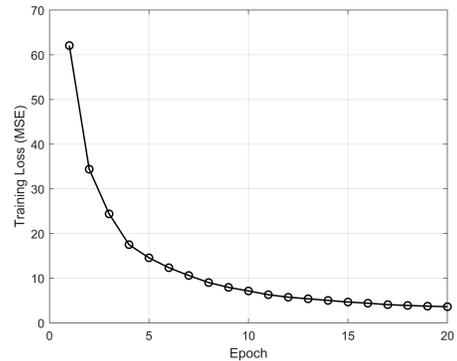

Figure 4: Training loss of deep CNN model on real vibroseis data experiment

1487 to generate train data and labels, and then we cut the train data and labels into patches of size 64*64 to train the model. The network has been trained for 20 epochs, the loss of deep CNN on train set is shown as Figure 4 . After the training process, we apply the trained deep CNN model to each gather in the real vibroseis data. Figure 5 shows No.1 to 252 traces of one seismic line in the vibroseis data (not in train set) before and after deep CNN deringing respectively in (a) and (b). To show the improvement for first-break

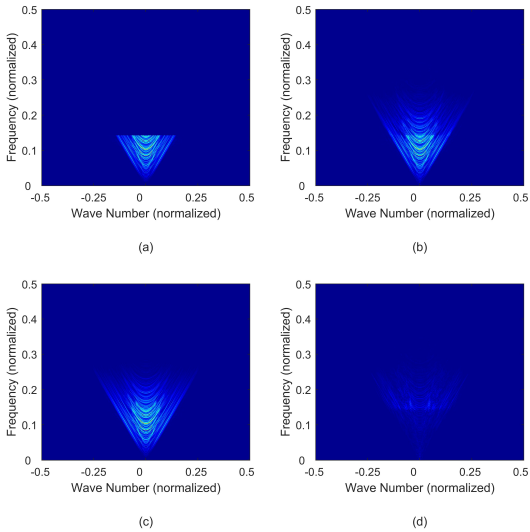

Figure 3: Comparison of synthetic data before and after deringing in f-K domain. (a) shows the f-K domain of band-pass filtered ringing synthetic data; (b) shows the f-K domain of deringed result; (c) shows the f-K domain of original synthetic data without ringing effect. (d) is the f-K domain of the difference between real data and our deringed result.

In the experiment with real vibroseis data, we use 8 gathers out of the real vibroseis data with a total number of gathers

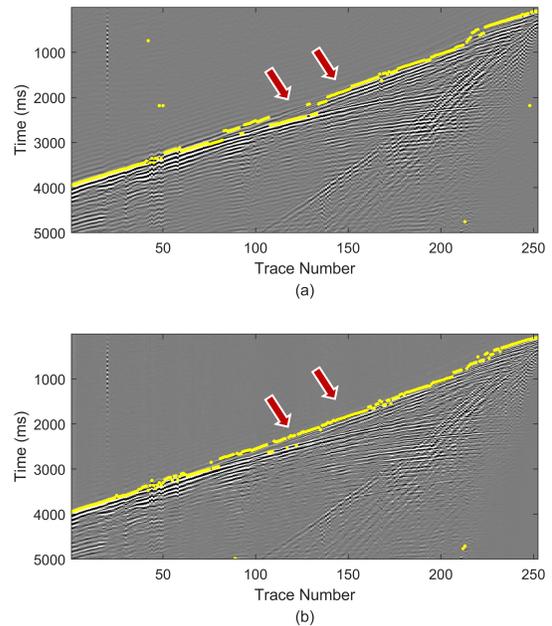

Figure 5: Automatic first-break picking performance on ringing data and deringed data (red arrows show the obvious ringing attenuation and first-break picking improvement)

- 3 -

picking, we apply the STA/LTA method to automatically pick the first-break trace by trace, and the first-break picked in each trace is marked with yellow dots. The results show that the first-break picked in the deep CNN deringed vibroseis data has more consistency and continuity than that of original vibroseis data with ringing effect.

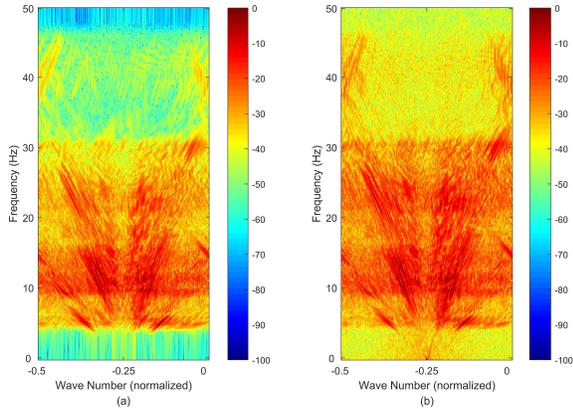

Figure 6: Comparison of real vibroseis data before and after deep CNN deringing in f-K domain (unit of values is dB). (a) shows the f-K domain before deringing, (b) shows the f-K domain after our deep CNN deringing approach.

Figure 6 compares the f-K domain between the ringing data and deringed data. The comparison shows that the frequency bandwidth is expanded both at high and low frequencies after applying our deep CNN deringing approach.

**Conclusions**

In this paper, a deep CNN model is proposed for ringing effect attenuation of vibroseis data. The deep CNN model has an end-to-end structure which can output the deringed seismic data directly. The training process of deep CNN model is conducted on the generated data and labels according to real vibroseis data. Both synthetic data experiment and real data experiment show that our model can attenuate the ringing effect of seismic data effectively by expanding the frequency bandwidth of ringing data. Automatic first break picking experiment using STA/LTA method shows that first break picking on deringed vibroseis data has more accuracy than that of original data with ringing effect.